\documentclass{article} 
\usepackage{iclr2016_conference,times}
\usepackage{hyperref}
\usepackage{url}
\usepackage{float}
\usepackage{graphicx}
\usepackage{subfig}

\title{sense2vec - a fast and accurate method \\ for word sense disambiguation in \\ neural word embeddings.}

\author{Andrew Trask \& Phil Michalak \& John Liu \\
Digital Reasoning Systems, Inc.\\
Nashville, TN 37212, USA \\
\texttt{\{andrew.trask,phil.michalak,john.liu\}@digitalreasoning.com} \\
}

\begin{document}

\maketitle

\begin{abstract}
Neural word representations have proven useful in Natural Language Processing (NLP) tasks due to their ability to efficiently model complex semantic and syntactic word relationships. However, most techniques model only one representation per word, despite the fact that a single word can have multiple meanings or "senses". Some techniques model words by using multiple vectors that are clustered based on context. However, recent neural approaches rarely focus on the application to a consuming NLP algorithm. Furthermore, the training process of recent word-sense models is expensive relative to single-sense embedding processes. This paper presents a novel approach which addresses these concerns by modeling multiple embeddings for each word based on supervised disambiguation, which provides a fast and accurate way for a consuming NLP model to select a sense-disambiguated embedding. We demonstrate that these embeddings can disambiguate both contrastive senses such as nominal and verbal senses as well as nuanced senses such as sarcasm. We further evaluate Part-of-Speech disambiguated embeddings on neural dependency parsing, yielding a greater than 8\% average error reduction in unlabeled attachment scores across 6 languages.
\end{abstract}

\section{Introduction}

NLP systems seek to automate the extraction of information from human language. A key challenge in this task is the complexity and sparsity in natural language, which leads to a phenomenon known as the curse of dimensionality. To overcome this, recent work has learned real valued, distributed representations for words using neural networks \citep{g.e.hintonj.l.mcclellandd.e.rumelhart1986,Bengio:2003:NPL:944919.944966,morin2005hierarchical,NIPS2008_3583}. These "neural language models" embed a vocabulary into a smaller dimensional linear space that models "the probability function for word sequences, expressed in terms of these representations" \citep{Bengio:2003:NPL:944919.944966}. The result is a vector-space model (VSM)  that represents word meanings with vectors that capture the semantic and syntactic information of words \citep{maas2010probabilistic}. These distributed representations model shades of meaning across their dimensions, allowing for multiple words to have multiple real-valued relationships encoded in a single vector \citep{liang2015semantics}. 

Various forms of distributed representations have shown to be useful for a wide variety of NLP tasks including Part-of-Speech tagging, Named Entity Recognition, Analogy/Similarity Querying, Transliteration, and Dependency Parsing \citep{DBLP:journals/corr/Al-RfouPS13,polyglotner,DBLP:journals/corr/abs-1301-3781,DBLP:journals/corr/MikolovLS13,chen-manning:2014:EMNLP2014}. Extensive research has been done to tune these embeddings to various tasks by incorporating features such as character (compositional) information, word order information, and multi-word (phrase) information \citep{ling-EtAl:2015:NAACL-HLT,DBLP:journals/corr/MikolovSCCD13,DBLP:journals/corr/ZhangZL15,DBLP:journals/corr/TraskGR15}.

Despite these advancements, most word embedding techniques share a common problem in that each word must encode all of its potential meanings into a single vector \citep{Huang:2012:IWR:2390524.2390645}. For words with multiple meanings (or "senses"), this creates a superposition in vector space where a vector takes on a mixture of its individual meanings. In this work, we will show that this superposition obfuscates the context specific meaning of a word and can have a negative effect on NLP classifiers leveraging the superposition as input data. Furthermore, we will show that disambiguating multiple word senses into separate embeddings alleviates this problem and the corresponding confusion to an NLP model.

\section{Related Work}

\subsection{Word2vec}

\citet{DBLP:journals/corr/abs-1301-3781} proposed two simple methods for learning continuous word embeddings using neural networks based on Skip-gram or Continuous-Bag-of-Word (CBOW) models and named it word2vec. Word vectors built from these methods map words to points in space that effectively encode semantic and syntactic meaning despite ignoring word order information. Furthermore, the word vectors exhibited certain algebraic relations, as exemplified by example: "v[man] - v[king] + v[queen] $\approx$ v[woman]". Subsequent work leveraging such neural word embeddings has proven to be effective on a variety of natural language modeling tasks \citep{DBLP:journals/corr/Al-RfouPS13,polyglotner,chen-manning:2014:EMNLP2014}.

\subsection{Wang2vec}

Because word embeddings in word2vec are insensitive to word order, they are suboptimal when used for syntactic tasks like POS tagging or dependency parsing. \citet{ling-EtAl:2015:NAACL-HLT} proposed modifications to word2vec that incorporated word order. Consisting of structured skip-gram and continuous window methods that are together termed wang2vec, these models demonstrate significant ability to model syntactic representations. They come, however, at the cost of computation speed. Furthermore, because words have a single vector representation in wang2vec, the method is unable to model polysemic words with multiple meanings. For instance, the word "work" in the sentence "We saw her work" can be either a verb or noun depending on the broader context in surrounding this sentence. This technique encodes the co-occurrence statistics for each sense of a word into one or more fixed dimensional embeddings, generating embeddings that model multiple uses of a word.

\subsection{Statistical Multi-Prototype Vector-Space Models of Word Meaning}

Perhaps a seminal work to vector-space word-sense disambiguation, the approach by \citet{Reisinger:2010:MVM:1857999.1858012} creates a vector-space model that encodes multiple meanings for words by first clustering the contexts in which a word appears. Once the contexts are clustered, several prototype vectors can be initialized by averaging the statistically generated vectors for each word in the cluster. This process of computing clusters and creating embeddings based on a vector for each cluster has become the canonical strategy for word-sense disambiguation in vector spaces. However, this approach presents no strategy for the context specific selection of potentially many vectors for use in an NLP classifier.

\subsection{Clustering Weighted Average Context Embeddings}

Our technique is inspired by the work of \citet{Huang:2012:IWR:2390524.2390645}, which uses a multi-prototype neural vector-space model that clusters contexts to generate prototypes. Unlike \citet{Reisinger:2010:MVM:1857999.1858012}, the context embeddings are generated by a neural network in the following way: given a pre-trained word embedding model, each context embedding is generated by computing a weighted sum of the words in the context (weighted by tf-idf). Then, for each term, the associated context embeddings are clustered. The clusters are used to re-label each occurrence of each word in the corpus. Once these terms have been re-labeled with the cluster's number, a new word model is trained on the labeled embeddings (with a different vector for each) generating the word-sense embeddings.

In addition to the selection problem and clustering overhead described in the previous subsection, this model also suffers from the need to train neural word embeddings twice, which is a very expensive endeavor.

\subsection{Clustering Convolutional Context Embeddings}

Recent work has explored leveraging convolutional approaches to modeling the context embeddings that are clustered into word prototypes. Unlike previous approaches, \citet{chen-EtAl:2015:ACL-IJCNLP1} selects the number of word clusters for each word based on the number of definitions for a word in the WordNet Gloss (as opposed to other approaches that commonly pick a fixed number of clusters). A variant on the MSSG model of \citet{DBLP:journals/corr/NeelakantanSPM15}, this work uses the WordNet Glosses dataset and convolutional embeddings to initialize the word prototypes. 

In addition to the selection problem, clustering overhead, and the need to train neural embeddings multiple times, this higher-quality model is somewhat limited by the vocabulary present in the English WordNet resource. Furthermore, the majority of the WordNet’s relations connect words from the same Part-of-Speech (POS). "Thus, WordNet really consists of four sub-nets, one each for nouns, verbs, adjectives and adverbs, with few cross-POS pointers."\footnote{https://wordnet.princeton.edu/}

\section{The sense2vec Model}

We expand on the work of \citet{Huang:2012:IWR:2390524.2390645} by leveraging supervised NLP labels instead of unsupervised clusters to determine a particular word instance's sense. This eliminates the need to train embeddings multiple times, eliminates the need for a clustering step, and creates an efficient method by which a supervised classifier may consume the appropriate word-sense embedding.

\begin{figure}[H]
\begin{center}  
\includegraphics[height=1.5in]{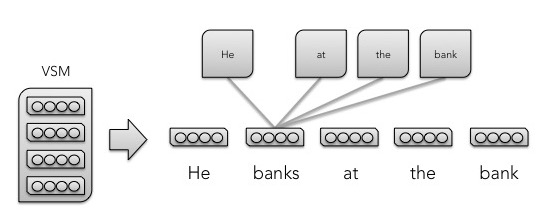}  
\caption{\small \sl A graphical representation of wang2vec.\label{fig:Droll}}  
\end{center}  
\end{figure}

\begin{figure}[H]
\begin{center}  
\includegraphics[height=2in]{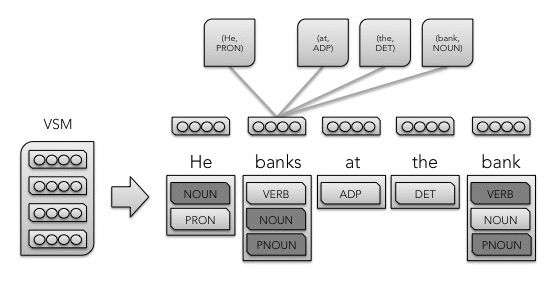}  
\caption{\small \sl A graphical representation of sense2vec. \label{fig:Stupendous}}  
\end{center}  
\end{figure}  

Given a labeled corpus (either by hand or by a model) with one or more labels per word, the sense2vec model first counts the number of uses (where a unique word maps set of one or more labels/uses) of each word and generates a random "sense embedding" for each use. A model is then trained using either the CBOW, Skip-gram, or Structured Skip-gram model configurations. Instead of predicting a token given surrounding tokens, this model predicts a word sense given surrounding senses. 

\subsection{Subjective Evaluation - Subjective Baseline}

For subjective evaluation of these word embeddings, we trained models using several datasets for comparison. First, we trained using Word2vec's Continuous Bag of Words \footnote{command line params: -size 500 -window 10 -negative 10 -hs 0 -sample 1e-5 -iter 3 -min-count 10} approach on the large unlabeled corpus used for the Google Word Analogy Task \footnote{the data.txt file generated from http://word2vec.googlecode.com/svn/trunk/demo-train-big-model-v1.sh}. Several word embeddings and their closest terms measured by cosine similarity are displayed in Table 1 below.

\begin{table}[ht]
\centering
\caption{Single-sense Baseline Cosine Similarities}
\label{my-label}
\begin{tabular}{|cc|cc|ll|ll|ll|}
\hline
\textbf{bank}                & \textbf{1.0}              & \textbf{apple}              & \textbf{1.0}              & \textbf{so} & \textbf{1.0} & \textbf{bad} & \textbf{1.0} & \textbf{perfect} & \textbf{1.0} \\ \hline
banks                        & .718                      & iphone                      & .687                      & but         & .879         & good         & .727         & perfection            & .681         \\
banking                      & .672                      & ipad                        & .649                      & it          & .858         & worse        & .718         & perfectly           & .670         \\
hsbc                         & .599                      & microsoft                   & .603                      & if          & .842         & lousy        & .717         & ideal         & .644         \\
citibank                     & .586                      & ipod                        & .595                      & even        & .833         & stupid       & .710         & flawless          & .637         \\
lender                       & .566                      & imac                        & .594                      & do          & .831         & horrible     & .703         & good           & .622         \\
\multicolumn{1}{|l}{lending} & \multicolumn{1}{l|}{.559} & \multicolumn{1}{l}{iphones} & \multicolumn{1}{l|}{.578} & just        & .808         & awful        & .697         & always      & .572        \\ \hline
\end{tabular}
\end{table}

In this table, observe that the "bank" column is similar to proper nouns ("hsbc", "citibank"), verbs ("lending","banking"), and nouns ("banks","lender"). This is because the term "bank" is used in 3 different ways, as a proper noun, verb, and noun. This embedding for "bank" has modeled a mixture of these three meanings. "apple", "so", "bad", and "perfect" can also have a mixture of meanings. In some cases, such as "apple", one interpretation of the word is completely ignored (apple the fruit). In the case of "so", there is also an interjection sense of "so" that is not well represented in the vector space. 

\subsection{Subjective Evaluation - Part-of-Speech Disambiguation}

For Part-of-Speech disambiguation, we labeled the dataset from section 3.1 with Part-of-Speech tags using the Polyglot Universal Dependency Part-of-Speech tagger of \citet{DBLP:journals/corr/Al-RfouPS13} and trained sense2vec with identical parameters as section 3.1. In table 2, we see that this method has successfully disambiguated the difference between the noun "apple" referring to the fruit and the proper noun "apple" referring to the company. In table 3, we see that all three uses of the word "bank" have been disambiguated by their respective parts of speech, and in table 4, nuanced senses of the word "so" have also been disambiguated.

\begin{table}[h]
\centering
\caption{Part-of-Speech Cosine Similarities for the Word: apple}
\label{my-label5}
\begin{tabular}{|ccc|ccc|}
\hline
\textbf{apple} & \textbf{NOUN} & \textbf{1.0} & \textbf{apple}              & \textbf{PROPN} & \textbf{1.0} \\ \hline
apples         & NOUN          & .639     & microsoft                   & PROPN          & .603     \\
pear           & NOUN          & .581     & iphone                      & NOUN           & .591     \\
peach          & NOUN          & .579     & ipad                        & NOUN           & .586     \\
blueberry      & NOUN          & .570     & samsung                     & PROPN          & .572     \\
almond         & NOUN          & .541     & blackberry                  & PROPN          & .564     \\ \hline
\end{tabular}
\end{table}

\begin{table}[ht]
\centering
\caption{Part-of-Speech Cosine Similarities for the Word: bank}
\label{my-label2}
\begin{tabular}{|ccc|ccc|ccc|}
\hline
\textbf{bank} & \textbf{NOUN} & \textbf{1.0} & \textbf{bank} & \textbf{PROPN} & \textbf{1.0} & \textbf{bank} & \textbf{VERB} & \textbf{1.0} \\ \hline
banks         & NOUN          & .786     & bank          & NOUN           & .570     & gamble        & VERB          & .533     \\
banking       & NOUN          & .629     & hsbc          & PROPN          & .536     & earn          & VERB          & .485     \\
lender        & NOUN          & .619     & citibank      & PROPN          & .523     & invest        & VERB          & .470     \\
bank          & PROPN         & .570     & wachovia      & PROPN          & .503     & reinvest      & VERB          & .466     \\
ubs           & PROPN         & .535     & grindlays     & PROPN          & .492     & donate        & VERB          & .466     \\ \hline
\end{tabular}
\end{table}

\begin{table}[H]
\centering
\caption{Part-of-Speech Cosine Similarities for the Word: so}
\label{my-label3}
\begin{tabular}{|ccc|ccc|ccc|}
\hline
\textbf{so} & \textbf{INTJ} & \textbf{1.0} & \textbf{so} & \textbf{ADV} & \textbf{1.0} & \textbf{so}          & \textbf{ADJ} & \textbf{1.0}      \\ \hline
now         & INTJ          & .527     & too         & ADV          & .753     & poved       & ADJ & .588 \\
obviously   & INTJ          & .520     & but         & CONJ         & .752     & condemnable & ADJ & .584 \\
basically   & INTJ          & .513     & because     & SCONJ        & .720     & disputable  & ADJ & .578 \\
okay        & INTJ          & .505     & but         & ADV          & .694     & disapprove  & ADJ & .559 \\
actually    & INTJ          & .503     & really      & ADV          & .671     & contestable & ADJ & .558 \\ \hline
\end{tabular}
\end{table}

\subsection{Subjective Evaluation - Sentiment Disambiguation}

For Sentiment disambiguation, the IMDB labeled training corpus was labeled with Part-of-Speech tags using the Polyglot Part-of-Speech tagger from \citet{DBLP:journals/corr/Al-RfouPS13}. Adjectives were then labeled with the positive or negative sentiment associated with each comment. A CBOW sense2vec model was then trained on the resulting dataset, disambiguating between both Part-of-Speech and Sentiment (for adjectives). 

Table 5 shows the difference between the positive and negative vectors for the word "bad". The negative vector is most similar to word indicating the classical meaning of bad (including the negative version of "good", e.g. "good grief!"). The positive "bad" vector denotes a tone of sarcasm, most closely relating to the positive sense of "good" (e.g. "good job!"). 

\begin{table}[H]
\centering
\caption{Sentiment Cosine Similarities for the Word: bad}
\label{my-label4}
\begin{tabular}{|ccc|ccc|}
\hline
\textbf{bad} & \textbf{NEG} & \textbf{1.0} & \textbf{bad} & \textbf{POS} & \textbf{1.0} \\ \hline
terrible     & NEG          & .905        & good         & POS          & .753        \\
horrible     & NEG          & .872        & wrong        & POS          & .752        \\
awful        & NEG          & .870        & funny        & POS          & .720        \\
good         & NEG          & .863        & great        & POS          & .694        \\
stupid       & NEG          & .845        & weird        & POS          & .671        \\ \hline
\end{tabular}
\end{table}

Table 6 shows the positive and negative senses of the word "perfect". The positive version of the word clusters most closely with words indicating excellence. The positive version clusters with the more sarcastic interpretation.

\begin{table}[H]
\centering
\caption{Sentiment Cosine Similarities for the Word: perfect}
\label{my-label7}
\begin{tabular}{|ccc|ccc|}
\hline
\textbf{perfect} & \textbf{NEG} & \textbf{1.0} & \textbf{perfect} & \textbf{POS} & \textbf{1.0} \\ \hline
real       & NEG & 0.682 & wonderful      & POS & 0.843  \\
unfortunate       & NEG & 0.680 & brilliant      & POS & 0.842  \\
serious   & NEG & 0.673 & incredible        & POS & 0.840  \\
complete & NEG & 0.673 & fantastic      & POS & 0.839  \\
ordinary      & NEG & 0.673 & great      & POS & 0.823  \\
typical        & NEG & 0.661 & excellent      & POS & 0.822  \\
misguided       & NEG & 0.650 & amazing & POS & 0.814  \\ \hline
\end{tabular}
\end{table}

\section{Named Entity Resolution}

To evaluate the embeddings when disambiguating on named entity resolution (NER), we labeled the standard word2vec dataset from section 3.2 with named entity labels. This demonstrated how sense2vec can also disambiguate between multi-word sequences of text as well as single word sequences of text. Below, we see that the word "Washington" is disambiguated with both a PERSON and a GPE sense of the word. Furthermore, we see that Hillary Clinton is very similar to titles that she has held within the time span of the dataset.

\begin{table}[H]
\centering
\caption{Disambiguation for the word: Washington}
\label{my-label8}
\begin{tabular}{|ccc|ccc|}
\hline
George\_Washington     & PERSON\_NAME          & .656        & Washington\_D         & GPE          & .665        \\
Henry\_Knox     & PERSON\_NAME          & .624        & Washington\_DC        & GPE          & .591       \\
Philip\_Schuyler        & PERSON\_NAME          & .618        & Seattle        & GPE          & .559        \\
Nathanael\_Greene         & PERSON\_NAME          & .613        & Warsaw\_Embassy        & GPE          & .524        \\
Benjamin\_Lincoln         & PERSON\_NAME          & .602        & Wash        & GPE          & .516        \\
William\_Howe       & PERSON\_NAME          & .591        & Maryland        & GPE          & .507        \\ \hline
\end{tabular}
\end{table}

\begin{table}[H]
\centering
\caption{Entity resolution for the term: Hillary Clinton}
\label{my-label9}
\begin{tabular}{|ccc|}
\hline
Secretary\_of\_State & TITLE & 0.661 \\
Senator & TITLE & 0.613 \\
Senate & ORG\_NAME & 0.564 \\
Chief & TITLE & 0.555 \\
White\_House & ORG\_NAME & 0.564 \\
Congress & ORG\_NAME & 0.547 \\ \hline
\end{tabular}
\end{table}

\section{Neural Dependency Parsing}

To quantitatively evaluate disambiguated sense embeddings relative to the current standard, we compared sense2vec embeddings and wang2vec embeddings on neural syntactic dependency parsing tasks in six languages. First, we trained two sets of embeddings on the Bulgarian, German, English, French, Italian, and Swedish Wikipedia datasets from the Polyglot website\footnote{https://sites.google.com/site/rmyeid/projects/polyglot}. The baseline embeddings were trained without any Part-of-Speech disambiguation using the structured skip-gram approach of \citet{ling-EtAl:2015:NAACL-HLT}. For each language, the sense2vec embeddings were trained by disambiguating terms using the language specific Polyglot Part-of-Speech tagger of \citet{DBLP:journals/corr/Al-RfouPS13}, and embedded in the same structured skip-gram approach. Both were trained using identical parametrization \footnote{command line params: -size 50 -window 5 -negative 10 -hs 0 -sample 1e-4 -iter 5 -cap 0}. 

Each of these embeddings was used to train a dependency parse model using the parser outlined in \citep{chen-manning:2014:EMNLP2014}. All were trained on the the respective language's Universal Dependencies treebank. The standard splits were used.\footnote{The German, French, and Italian treebanks had occasional tokens that both spanned multiple indices and overlapped with the index of the previous and following token (ex. 0, 0-1, 1,...), a property which is incompatible with the \citep{chen-manning:2014:EMNLP2014} parser. These tokens were removed. If their removal created a malformed tree, the sentence was removed automatically by the parser and logged accordingly.} For the parser trained on the sense2vec emeddings, the POS specific embedding was used as the input. The Part-of-Speech label was determined using the gold-standard POS tags from the treebank. It should be noted that the parser of \citep{chen-manning:2014:EMNLP2014} uses trained Part-of-Speech embeddings as input which are indexed based on gold-standard POS tags. Thus, differences in quality between parsers trained on the two embedding styles are due to clarity in the word embeddings as opposed to the addition of Part-of-Speech information because both model styles train on gold standard POS information. For each language, the Unlabeled Attachment Scores are outlined in Table 7.

\begin{table}[H]
\centering
\caption{Unlabeled Attachment Scores and Percent Error Reductions}
\label{my-label6}
\begin{tabular}{|cc|cccccc|c|}
\hline
        & \textbf{Set}      & \textbf{Bulgarian} & \textbf{German} & \textbf{English} & \textbf{French} & \textbf{Italian} & \textbf{Swedish} & \textbf{Mean} \\ \hline
        & Dev      & 90.03     & 68.86  & 85.02   & 73.82  & 84.99   & 78.94   & 80.28          \\
wang & Test*    & 90.17     & 60.25  & 83.61   & 70.10  & 84.99   & 82.47   &    78.60      \\
        & Test     & 90.39     & 60.54  & 83.88   & 70.53  & 85.45   & 82.51   & 78.88         \\ \hline
        & Dev      & 90.69     & 72.61  & 86.10   & 75.43  & 85.57   & 81.21   & 81.94         \\
sense & Test*    & 90.41     & 64.17  & 85.48   & 71.66  & 86.13   & 84.44   &   80.38       \\
        & Test     & 90.86     & 64.43  & 85.93   & 72.16  & 86.18   & 84.60   & 80.69          \\ \hline
        & Dev      & 7.05\%    & 13.69\% & 7.76\%  & 6.56\% & 3.98\%  & 12.06\%  & \textbf{8.52\%}   \\
Error   & Test     & 2.47\%    & 10.95\% & 12.82\%  & 5.50\% & 8.21\%  & 12.71\%  & \textbf{8.78\%}   \\
Margin  & Abs. & 5.17\%    & 10.93\% & 14.54\%  & 5.86\% & 5.32\%  & 13.58\%  & \textbf{9.23\%}   \\
        & Avg.  & 4.76\%    & 12.32\% & 10.29\%  & 6.03\% & 6.09\%  & 12.39\%  &          \\ \hline
\end{tabular}
\end{table}

The "Error Margin" section of table 7 describes the percentage reduction in error for each language. Disambiguating based on Part-of-Speech using sense2vec reduced the error in all six languages with an average reduction greater than 8\%.

\section{Conclusion and Future Work}

In this work, we have proposed a new model for word sense disambiguation that uses supervised NLP labeling to disambiguate between word senses. Much like previous models, it leverages a form of context clustering to disambiguate the use of a term. However, instead of using unsupervised clustering methods, our approach clusters using supervised labels which can analyze a specific word's context and assign a label. This significantly reduces the computational overhead of word-sense modeling and provides a natural mechanism for other NLP tasks to select the appropriate sense embedding. Furthermore, we show that disambiguated embeddings can increase the accuracy of syntactic dependency parsing in a variety of languages. Future work will explore how disambiguated embeddings perform using other varieties of supervised labels and consuming NLP tasks.

\bibliography{iclr2016_conference}
\bibliographystyle{iclr2016_conference}

\end{document}